\documentclass{article} 
\usepackage{colm2024_conference}

\usepackage{booktabs}
\usepackage{graphicx}
\usepackage{enumitem}
\usepackage{wrapfig}
\usepackage{algorithm}
\usepackage{algpseudocode}
\usepackage{natbib}
\usepackage{makecell}
\usepackage{booktabs}
\usepackage{array}
\usepackage{amsmath}
\usepackage{amssymb}
\usepackage{amsfonts}
\usepackage{threeparttable}
\usepackage{multirow}
\usepackage{verbatim}
\usepackage{caption}
\usepackage{placeins}
\usepackage{longtable}
\usepackage{ragged2e}      
\usepackage{array}         
\usepackage{supertabular}
\usepackage{CJKutf8}
\usepackage{array} 
\usepackage[utf8]{inputenc} 
\usepackage[T1]{fontenc}
\usepackage[french,vietnamese,mongolian,greek,english]{babel}
\usepackage{pifont}
\usepackage{xcolor}
\usepackage{enumitem}
\usepackage{tablefootnote}
\usepackage{xspace}
\usepackage{textcomp}
\usepackage{makecell}
\usepackage{lscape} 
\usepackage{siunitx}
\usepackage{listings}
\usepackage{xcolor}
\usepackage{colortbl}
\definecolor{kuaishoublue}{HTML}{6D9EEB}
\hypersetup{
    colorlinks=true,   
}
\definecolor{dt}{gray}{0.7}
\usepackage{pifont}       
\usepackage{bbding}       
\usepackage{fontawesome}
\newcolumntype{L}[1]{>{\raggedright\arraybackslash}m{#1}}

\usepackage{scrextend}

\usepackage{tgpagella}
\usepackage{latexsym}
\usepackage[T1]{fontenc}
\usepackage{microtype}
\definecolor{mydarkblue}{rgb}{0,0.08,0.45}
\definecolor{citecolor}{HTML}{0071BC}
\usepackage{url}            
\usepackage{nicefrac}       
\usepackage{changepage}
\usepackage{xargs}          
\usepackage{wrapfig,lipsum,booktabs}
\usepackage{longtable}
\usepackage{subcaption}
\usepackage{endnotes}

\usepackage{fancyvrb}
\usepackage{fvextra}
\usepackage{pgfplots}
\usetikzlibrary{pgfplots.groupplots}
\pgfplotsset{compat=1.3}
\usepackage{tikz}
\usetikzlibrary{patterns}

\usepackage[most]{tcolorbox}
\usepackage[capitalize,noabbrev]{cleveref}
\crefname{section}{Section}{\S\S}
\Crefname{section}{Section}{\S\S}
\crefname{table}{Table}{Tables}
\crefname{figure}{Figure}{Figures}
\crefname{algorithm}{Algorithm}{}
\crefname{equation}{eq.}{}
\crefname{appendix}{Appendix}{}
\crefformat{section}{Section #2#1#3}
\usepackage{multicol}
\usepackage{tcolorbox}

\usepackage{titlesec}
\titleformat*{\section}{\large\bfseries}
\AtBeginDocument{\setstretch{1.07}}

\title{Kling-Avatar: Grounding Multimodal Instructions for Cascaded Long-Duration Avatar Animation Synthesis}

\author{
\textbf{Yikang Ding}\thanks{Equal contribution. Sorted in alphabetical order by surname.}\quad \  
\textbf{Jiwen Liu}\footnotemark[1]\quad \    
\textbf{Wenyuan Zhang}\quad   
\textbf{Zekun Wang}\quad   
\textbf{Wentao Hu}
\\[0.1em]
\textbf{Liyuan Cui}\quad
\textbf{Mingming Lao}\quad
\textbf{Yingchao Shao}\quad
\textbf{Hui Liu}\quad
\textbf{Xiaohan Li}  
\\[0.1em]
\textbf{Ming Chen}\quad 
\textbf{Xiaoqiang Liu}\quad   
\textbf{Yu-Shen Liu}\quad
\textbf{Pengfei Wan} 
\\[0.75em]
Kling Team, Kuaishou Technology
\\[0.75em]
\textbf{Project Page: } \url{https://klingavatar.github.io/}}
\vspace{-0.3cm}

\begin{document}

\maketitle

\vspace{-0.5cm}
\begin{figure}[h]
    \centering
    \includegraphics[width=\linewidth]{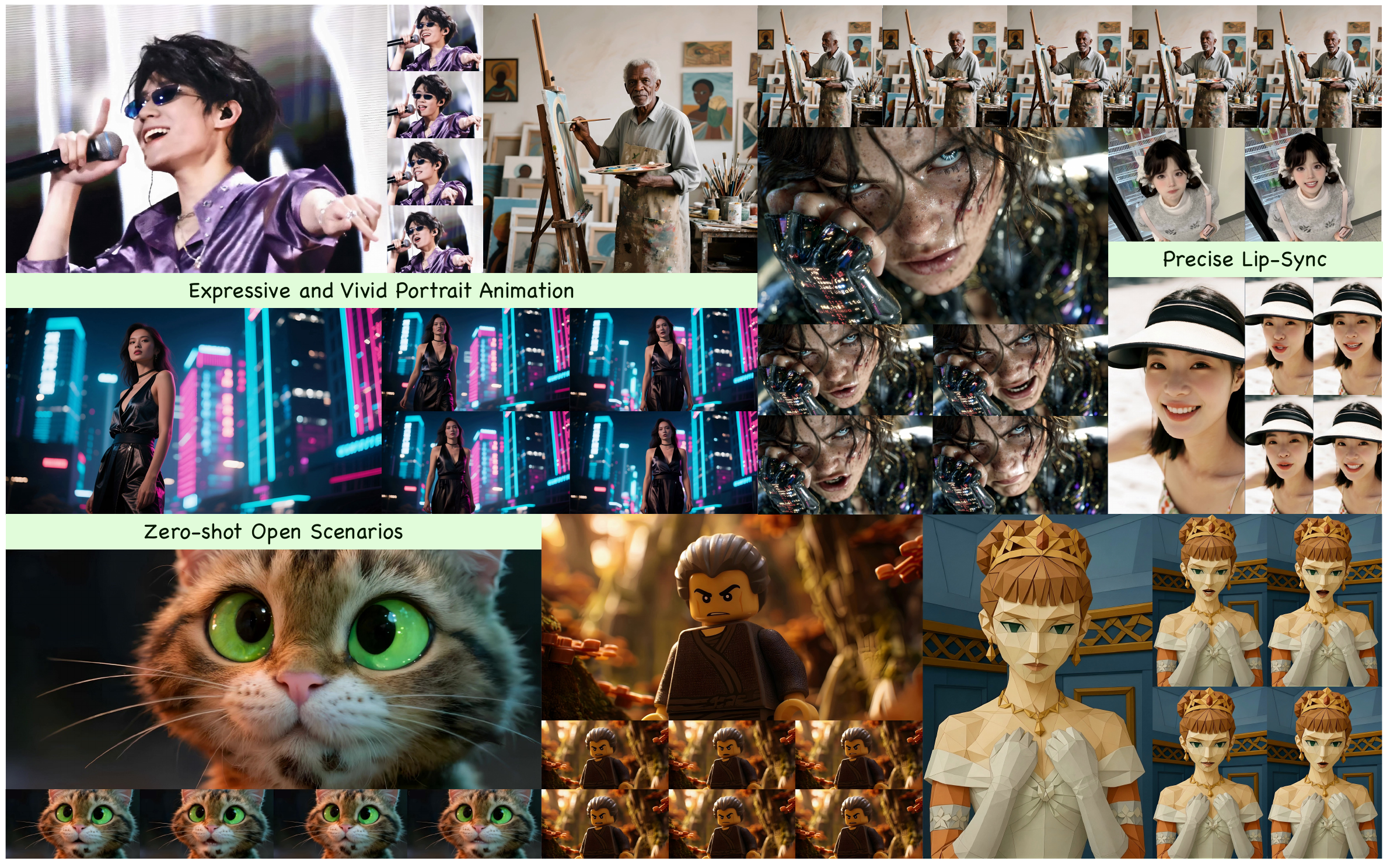}
    \vspace{-0.4cm}
    \caption{Conditioned on audio, image, and user prompts, \textbf{Kling-Avatar} generates high-fidelity portrait animations through instruction grounding and semantic planning. The results exhibit vivid emotions, rich actions, and precise lip synchronization, while also showing strong generalization to open scenarios such as anime, cartoons, and stylized characters.}
    \label{fig:teaser}
\end{figure}
\vspace{-0.0cm}
\begin{figure}[h]
    \centering
    \includegraphics[width=\linewidth]{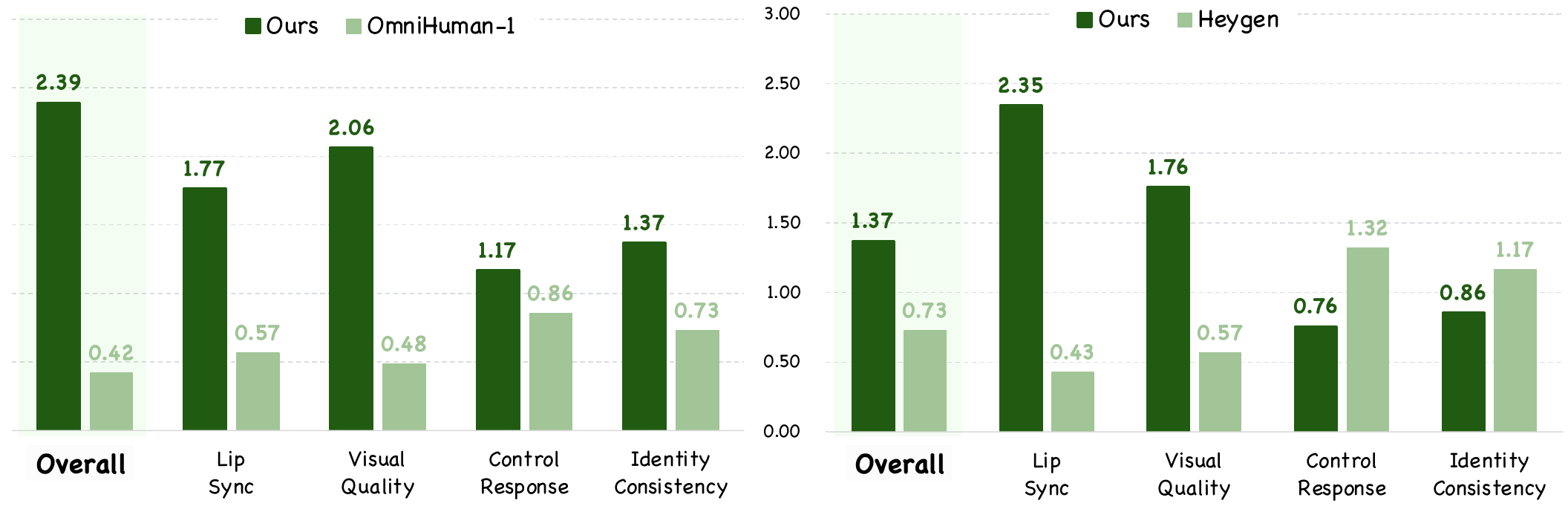}
    \vspace{-0.4cm}
    \caption{Benchmark performance of \textbf{Kling-Avatar} against its counterparts in terms of GSB metrics. We achieve superior performance on the overall metric as well as across most of sub-dimensions.}
    \label{fig:teaser-table}
\end{figure}

\clearpage

\begin{abstract}
Recent advances in audio-driven avatar video generation have significantly enhanced audio-visual realism. However, existing methods treat instruction conditioning merely as low-level tracking driven by acoustic or visual cues, without modeling the communicative purpose conveyed by the instructions. This limitation compromises their narrative coherence and character expressiveness.
To bridge this gap, we introduce Kling-Avatar, a novel cascaded framework that unifies multimodal instruction understanding with photorealistic portrait generation. Our approach adopts a two-stage pipeline. In the first stage, we design a multimodal large language model (MLLM) director that produces a blueprint video conditioned on diverse instruction signals, thereby governing high-level semantics such as character motion and emotions. In the second stage, guided by blueprint keyframes, we generate multiple sub-clips in parallel using a first-last frame strategy. 
This global-to-local framework preserves fine-grained details while faithfully encoding the high-level intent behind multimodal instructions. Our parallel architecture also enables fast and stable generation of long-duration videos, making it suitable for real-world applications such as digital human livestreaming and vlogging. To comprehensively evaluate our method, we construct a benchmark of 375 curated samples covering diverse instructions and challenging scenarios. Extensive experiments demonstrate that Kling-Avatar is capable of generating vivid, fluent, long-duration videos at up to 1080p and 48 fps, achieving superior performance in lip synchronization accuracy, emotion and dynamic expressiveness, instruction controllability, identity preservation, and cross-domain generalization. These results establish Kling-Avatar as a new benchmark for semantically grounded, high-fidelity audio-driven avatar synthesis.
Demonstration videos are available in our project page: \url{https://klingavatar.github.io/}.

\end{abstract}

\section{Introduction}

Avatar animation synthesis translates multimodal references into temporally coherent facial expressions, lip movements and body gestures, enabling interactions with machines that feel conversational and embodied. As a communicative medium, a speaking avatar can convey intent and affect with high fidelity, turning abstract ideas into vivid, situated performances that maintain user attention and improve comprehension. This capability opens broad opportunities across virtual assistants, education, media content creation, and immersive telepresence. Building such avatars requires models that couple realism with fine-grained controllability and reliable synchronization, which define the core challenge and motivate the approach developed in this work.

Recently, Video Diffusion Transformers (DiT)~\citep{chen2025echomimic,cui2025hallo3,jiang2024loopy,tian2024emo,peng2025omnisync,wei2025mocha} have emerged as a general paradigm for generating visually compelling content conditioned on multimodal signals such as images, speech, and prompts. Prior work has advanced precise facial expression and lip synchronization~\citep{jiang2024loopy,fei2025skyreels,tian2024emo}, coordinated body motion~\citep{wang2025fantasytalking,gan2025omniavatar}, and data scaling~\citep{lin2025omnihuman1,jiang2025omnihuman1.5}. However, these advances remain insufficient for highly realistic portrait synthesis, as we want the system not only to hear and to read but also to understand these inputs, so that it can produce natural and empathetic videos aligned with user intents. Without such understanding, current approaches often treat each conditional signal independently and capture only shallow correlations, which leads to semantic conflicts across modalities and affect. For example, an avatar may sing a sorrowful song while smiling, which is visually polished yet inconsistent with human expectations. In addition, existing approaches often rely on motion frames for video continuation, which poses significant challenges for maintaining consistency and stability in long-duration generation.

To bridge this gap, we introduce Kling-Avatar, a novel cascaded framework for portrait animation that faithfully follows multimodal instructions and synthesizes high-quality and long-duration avatar videos. Drawing inspiration from the capabilities of unifying understanding and generation of multimodal large language models(MLLMs)~\citep{team2023gemini,xu2025qwen2.5omni,hong2025glm4.1}, we design an MLLM Director that consolidates multimodal instructions into a structured storyline. This storyline encodes high-level plans such as scene layout, camera positioning, character motions, as well as implicit emotions and atmosphere, ensuring that the generated content aligns with the intended narrative arc and expressive trajectory. A blueprint video is first generated conditioned on the global script, followed by multiple sub-clips generation in parallel conditioned on the blueprint keyframes. The MLLM Director continuously provides fine-grained guidance based on multimodal context, ensuring local dynamics and visual details. By densely selecting anchor frames and enabling parallel generation, our cascaded framework supports fast and stable synthesis of arbitrarily long videos, offering a promising solution for long-term digital human video generation. 

For data preparation, we collect a dataset covering diverse scenarios such as dialogues, films, and speeches. To ensure data quality, we employ a series of expert models for rigorous filtering, including mouth-clarity recognition, stage-cut detection, audio-lip synchronization checking, and video quality scoring. To validate the effectiveness of our method, we construct a unique benchmark containing 375 “reference frame–audio” pairs. For these cases, we carefully design challenging instructions that include images from diverse categories, audio spanning different languages and speech rates, and text prompts with explicit control over emotion and dynamics. This benchmark builds up a comprehensive evaluation of different methods across multiple dimensions. 
We highlight representative generation results in Fig.~\ref{fig:teaser}, where Kling-Avatar produces expressive, vivid, long-duration portrait animations with rich emotions and dynamics, while maintaining strong generalization to open-domain scenarios. As indicated by the comparisons in Tab.~\ref{fig:teaser-table} against leading competitors OmniHuman-1~\citep{lin2025omnihuman1} and HeyGen~\citep{heygen}, Kling-Avatar achieves superior performance in terms of lip synchronization accuracy, visual fidelity, instruction conditioned expressiveness, and identity preservation over long-duration generation. These results establish Kling-Avatar as the new benchmark for controllable, high-fidelity digital portrait animation synthesis.
We summarize our contributions as follows:

\begin{itemize}
    \item \textbf{MLLM Director with unified instruction grounding.} We introduce an MLLM Director that grounds multimodal instructions into unified global plan, providing a new perspective that lifts portrait video generation from tracking low-level cues to semantic and intent understanding.
    \item \textbf{Cascaded avatar animation synthesis framework.} We design a two-stage generation pipeline that first establishes high-level semantic guidance and then refines local dynamics, enabling long-duration video generation with coherent and expressive performances.
    \item \textbf{Curated data construction pipeline.} We develop a data filtering pipeline powered by expert models for quality control, and further construct a challenging benchmark to enable comprehensive evaluation of digital human generation systems.
    \item \textbf{High-fidelity performance and strong generalization.} Kling-Avatar produces state-of-the-art coherent and vivid portrait animations with precise lip synchronization, rich facial expressions and accurate response for multimodal instructions across diverse scenarios.
\end{itemize}

\section{Method}
\begin{figure}[t]
    \centering
    \includegraphics[width=\linewidth]{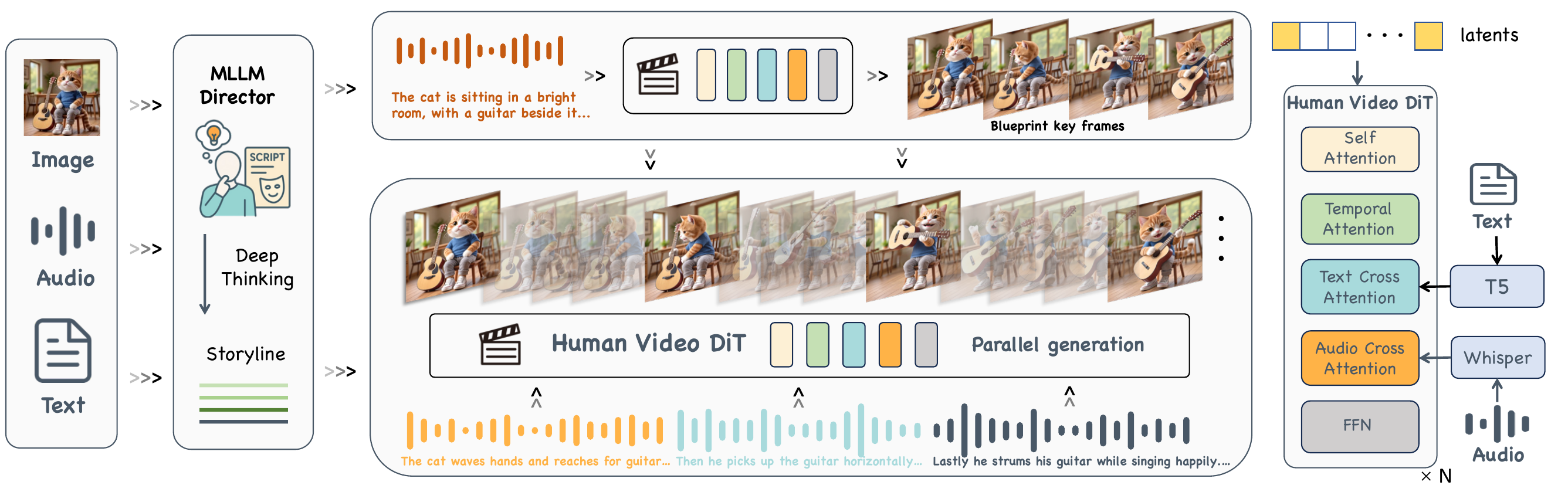}
    \caption{Illustration of Kling-Avatar’s cascaded generation pipeline. An MLLM Director first interprets multimodal instructions into high-level semantics and tells a storyline. Guided by this global planning, the first stage generates a blueprint video. In the second stage, keyframes are extracted from the blueprint and used as first–last frame conditions for parallel sub-clip generation, refining local details and dynamics to synthesize long-duration videos.}
    \label{fig:method-pipeline}
\end{figure}

Given a conditioning image, audio, and text prompt, Kling-Avatar aims to generate fluent and lifelike portrait animations with precise lip synchronization, accurate instruction following, and support for long-term extrapolation. As illustrated in Fig.~\ref{fig:method-pipeline}, our framework is a two-stage generation pipeline guided by an MLLM Director. In the following sections, we first present the motivation and implementation of using MLLMs for instruction grounding and control (Sec.~\ref{sec:mllm}). We then introduce the cascaded generation framework (Sec.~\ref{sec:cascaded}) for long-duration video synthesis, followed by our efforts in data construction for training and benchmarking (Sec.~\ref{sec:data}). Finally, we describe several key strategies for training and inference (Sec.~\ref{sec:train}).

\subsection{Grounding Multimodal Instructions with MLLMs}
\label{sec:mllm}
Current digital human video generation methods focus on conditioning strategies such as sliding windows or multi-scale injection, to better align input signals with the denoising diffusion process~\citep{gao2025wans2v,fei2025skyreels,wang2025fantasytalking}. However, this alignment is typically performed per modality, relying on local cues such as acoustic features or pixel structures, followed by shallow fusion at the generation stage. While effective at reproducing observable details, this paradigm lacks coordination across multimodal inputs, leading to semantic conflicts or impoverished camera language. For instance, when the input contains angry speech but the text imposes no such constraint, the emotion may be significantly weakened in the final output. 

To enable the model to truly understand the intent behind the instructions, drawing inspiration from multimodal large language models (MLLMs)~\citep{bai2025qwen2.5vl,hong2025glm4.1,qi2025quicksviewer}, we unify evidence from multimodal inputs into a shared semantic space, producing high-level control signals as a global planning for the generation process. 
Specifically, we use Qwen2.5-Omni~\citep{xu2025qwen2.5omni} to extract the transcription and emotion from audio as the audio caption, and Qwen2.5-VL~\citep{bai2025qwen2.5vl} to generate descriptions from images as the image caption. These captions are then combined with the user prompt and processed by our MLLM Director, to output a coherent storyline. We explicitly specify the storyline template for the MLLM Director using a three-shot in-context learning manner. This storyline, prioritized by user knowledge, audio, and image references, tells key elements such as character features, background layout, actions, visual style, camera planning, and emotional shifts. All of these elements are organized into a unified textual prompt, which is injected into the video diffusion model through a text cross-attention layer to generate a blueprint video.

\subsection{Cascaded Generation for Long-Duration Generation}
\label{sec:cascaded}

In the first stage, we generate a blueprint video that tells a storyline to reflect semantic user intent. The blueprint is then leveraged in the second stage to produce video sub-clips that refine local dynamics and visual details. To this end, we evenly segment the video according to the desired number of clips. Around each segmentation point, we then select a high-quality frame that preserves identity consistency, exhibits significant motion, avoids occlusion, and conveys expressive facial details. These frames serve as anchor keyframes for first-last-frame conditioned generation of adjacent sub-clips. During sub-clip synthesis, the MLLM Director decomposes the global storyline into temporally localized semantic plans. This localized narrative, combined with time-aligned audio conditioning, provide fine-grained guidance to ensure expressive coherence and visual consistency throughout the generated sequence. To avoid misalignment between anchor frames and the actual speech timing, we employ an audio-conditioned interpolation strategy to synthesize transition frames. This ensures precise frame synchronization with the input audio, enabling seamless and temporally coherent transitions across sub-clips.

The pipeline can be easily implemented in a parallel manner since the clips are generated independently. By increasing the anchor number, arbitrarily long videos can be generated with nearly the same runtime as producing a single clip. This cascaded framework with first-last-frame conditioned generation highlights our unique advantage in generating long-duration videos and provides a promising solution for downstream applications such as digital human podcasting, public speaking, and online education.

\subsection{Data Preparation}
\label{sec:data}

\textbf{Training Data.} We collect thousands of hours of audio–visual content from multiple sources, including publicly available datasets as well as self-collected videos such as film clips, speeches, monologues, interviews, and singing performances, covering a wide range of scenarios, linguistic styles and character dynamics. All videos are carefully processed with audio extraction and captioning. In our practice, we emphasize that data quality, rather than data scale, plays a decisive role in final performance: a smaller amount of high-quality talking segments proves more effective than indiscriminately enlarging the dataset with long-tail samples. To this end, we design a suite of expert models to classify and filter low-quality data along multiple dimensions: 
\begin{itemize}
    \item \textbf{Lip-clarity filtering.} We synthetically perturb mouth regions in high-quality talking-head videos to construct positive/negative pairs. A binary discriminator is trained to classify lip-region clarity and filter out videos with visually ambiguous or motion-blurred lip movements.
    \item \textbf{Temporal-continuity detection.} We manually assemble different video segments to build negative samples, paired with original clips as positives. We then train a temporal coherence discriminator, along with PySceneDetect~\citep{pyscenedetect}, to identify and remove discontinuous clips.
    \item \textbf{Audio–visual synchronization.} We employ SyncNet~\citep{chung2016syncnet} to assess frame-level audio-visual synchronization confidence scores, and discard videos that fall below calibrated thresholds.
    \item \textbf{Aesthetic quality assessment.} We adopt video aesthetic scoring methods~\citep{schuhmann2022Aesthetic} to evaluate visual composition and appeal. Only videos exceeding a calibrated quality threshold are included in the final training set.
\end{itemize}
After filtering the data using expert models, we further perform manual curation on the retained samples, ultimately assembling hundreds of hours of high-quality human portrait videos, which provide reliable supervision for training our model.

\textbf{Benchmark.} To comprehensively evaluate the performance of Kling-Avatar, we construct a challenging benchmark comprising 375 image–audio-prompt pairs. The dataset is carefully designed with the following composition:
\begin{itemize}
    \item \textbf{Images.} Reference images are sourced equally from real videos and AI-generated content. The set includes 340 human portraits of different races in both full-body and half-body formats, as well as 35 non-human cases covering cartoon, anime, and animal characters. Image resolutions span vertical, horizontal, and square formats, ranging from 480p to 1080p.
    \item \textbf{Audio.} Audio tracks are extracted from real videos and cover both speeches and songs. The collection includes 150 Chinese, 150 English, 35 Korean, and 40 Japanese samples, with clip lengths ranging from 8 seconds to 2 minutes. The audios span multiple speaking rates and expressive styles, ensuring diversity in linguistic and prosodic conditions.
    \item \textbf{Prompt.} Text prompts are manually annotated with diverse and explicit specifications on emotional expression, character actions, camera movements, and background layouts. Emotion categories include calm, excitement, confusion, sadness, surprise, and anger, each with multiple intensity levels. Camera instructions specify operations such as panning and zooming. Action descriptions encompass turning, raising hands, head shaking, and other expressive gestures, ensuring broad coverage of dynamic behaviors.
\end{itemize}
This benchmark establishes a demanding testbed for existing methods by requiring vivid and coherent portrait generation under complex multimodal instruction control.

\subsection{Training and Inference Strategy}
\label{sec:train}
\textbf{Training Strategy.} We design several training strategies to strengthen the alignment between lip movements and the corresponding speech. First, we adopt a sliding window scheme to inject audio features into the audio cross-attention layer, where each video token attends only to its temporally aligned audio tokens with a small padding, thereby reinforcing local phase consistency. Second, we employ DWPose~\citep{yang2023dwpose} to locate the mouth region and assign a higher weight to its diffusion denoising loss. Third, we randomly pad empty pixels around video frames during training to reduce the proportion of the face in the image, which encourages the model to remain robust under small-face and long-shot conditions. Finally, to preserve the text controllability of the base video generation model and concentrate on audio–visual interaction, we freeze the parameters of the text cross-attention layer during training, effectively preventing the base model from collapsing into overfitting the specific talking-head data. Collectively, these strategies substantially improve lip synchronization accuracy by enhancing visual-audio alignment.

\textbf{Inference Strategy.} Our first-last-frame conditioned parallel generation framework alleviates the identity drift problem that commonly arises in existing methods which rely on motion frames for long video continuation. To further improve identity consistency within each segment, we introduce a negative frame Classifier-Free Guidance (CFG) mechanism. Through statistical analysis, we find that identity drift artifacts typically manifest as texture distortions, blurring, exaggerated contrast and saturation, and color shifts. To counter this, we manually corrupt the reference image according to these observed patterns to simulate an enhanced identity drift. The degraded image is then used as a negative CFG signal to guide the denoising process toward identity-consistent directions. In addition, since no ground truth frames are available for mouth region masking during inference, we instead increase the audio cross-attention values to strengthen the lip-audio alignment.

\begin{table}[t]
  \vspace{-0.0cm}
  \centering
  \caption{Numerical evaluations on GSB metrics between our method and competitors.}
  \vspace{-0.1cm}
  \label{tab:gsb}
  \resizebox{\textwidth}{!}{
  \begin{tabular}{l|l|ccccc}
    \toprule
      Category & GSB & Overall & Lip Sync & Visual Quality & Control Response & ID Consistency \\
    \midrule
      \multirow{2}{*}{Overall} & Ours vs. OmniHuman & 2.39 & 1.77 & 2.06 & 1.17 & 1.37 \\
       & Ours vs. HeyGen & 1.37 & 2.35 & 1.76 & 0.76 & 0.86 \\
    \cmidrule{1-7}
      \multirow{2}{*}{Speech-En} & Ours vs. OmniHuman & 1.41 & 1.00 & 2.18 & 1.06 & 1.27 \\
       & Ours vs. HeyGen & 0.79 & 1.22 & 1.51 & 0.83 & 0.76 \\
    \cmidrule{1-7}
      \multirow{2}{*}{Speech-Ch} & Ours vs. OmniHuman & 4.53 & 3.90 & 2.44 & 1.13 & 1.47 \\
       & Ours vs. HeyGen & 1.22 & 2.26 & 1.93 & 0.79 & 0.82 \\
    \cmidrule{1-7}
      \multirow{2}{*}{Sing-En/Ch} & Ours vs. OmniHuman & 2.69 & 2.03 & 1.72 & 1.35 & 1.38 \\
       & Ours vs. HeyGen & 2.90 & 7.69 & 1.89 & 0.97 & 0.70 \\
  \bottomrule
    \end{tabular}
    }
\end{table}
\begin{figure}[t]
    \centering
    \includegraphics[width=\linewidth]{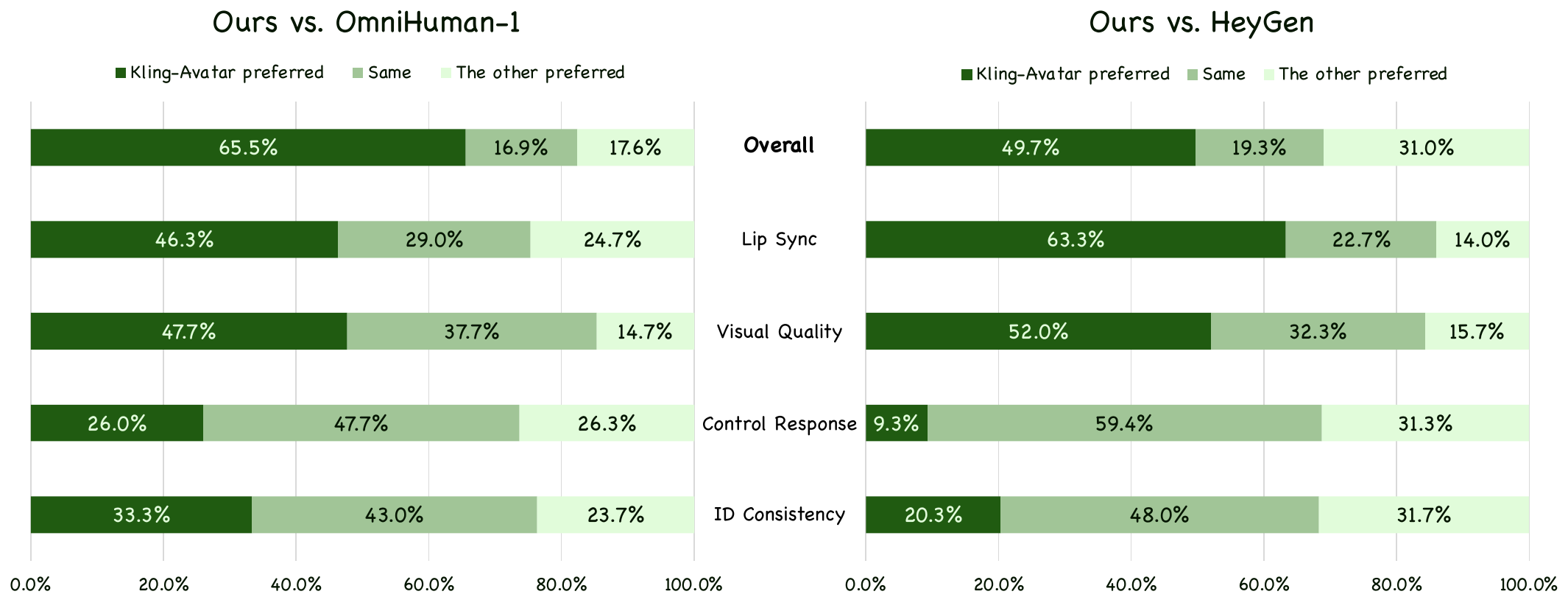}
    \vspace{-0.2cm}
    \caption{Overall GSB evaluation results on our benchmark across various dimensions against OmniHuman-1 and HeyGen.}
    \label{fig:visual-exp-gsb}
\end{figure}

\section{Experiments}

\subsection{Experimental Settings}

\textbf{Implementation Details.} Our implementation is based on a Video Diffusion Transformer architecture which was pretrained on a large-scale dataset. We extend it with an audio cross-attention layer to support audio-to-video generation. Audio features are extracted via a pre-trained Whisper encoder~\citep{radford2022whisper}, and text conditioning utilizes a T5 encoder~\citep{raffel2020t5}. The model is optimized using AdamW~\citep{loshchilov2017adamw} with a learning rate of 1e-5. During training, our framework supports arbitrary video resolutions ranging from 480p to 1080p, and at inference it produces fluent videos with up to 1080p at 48 fps.

\textbf{Evaluation Metrics.} We design a human preference–based subjective evaluation protocol as our primary metric, aiming to better reflect user-perceived semantics and aesthetic quality. For each sample in the benchmark, three participants independently provide a Good/Same/Bad (GSB) judgment by comparing the results of our method against baseline methods, and the final GSB label is determined by majority vote. We report (G+S)/(B+S) as the main metric, reflecting the proportion of cases where our method is judged as "better or not worse" than the baseline. In addition to the overall evaluation, we also conduct GSB assessments on four specific dimensions, including:
\begin{itemize}
    \item \textbf{Lip Synchronization.} Assesses the naturalness of lip movements, accuracy of audio–visual alignment, and plausibility of facial expressions.
    \item \textbf{Visual Quality.} Evaluates overall aesthetic appeal, structural coherence, and visual clarity of the generated video.
    \item \textbf{Control Response.} Examines whether emotions, actions, and camera movements in the generated video accurately reflect the textual instructions. Since OmniHuman-1 does not support prompt input, this metric is instead used to evaluate how effectively audio conditions control the body movements.  
    \item \textbf{Identity Consistency.} Measures how well the generated video preserves identity traits and dynamic characteristics that are consistent with the reference image.  
\end{itemize}
This GSB protocol provides a unified and intuitive framework of evaluating key aspects such as multimodal instruction following, avatar expressiveness, and visual coherence, which better reflects user subjective experience in real-world scenarios. We plan to incorporate additional objective metrics in the future to complement and extend our assessment.

\begin{figure}[t]
    \centering
    \includegraphics[width=\linewidth]{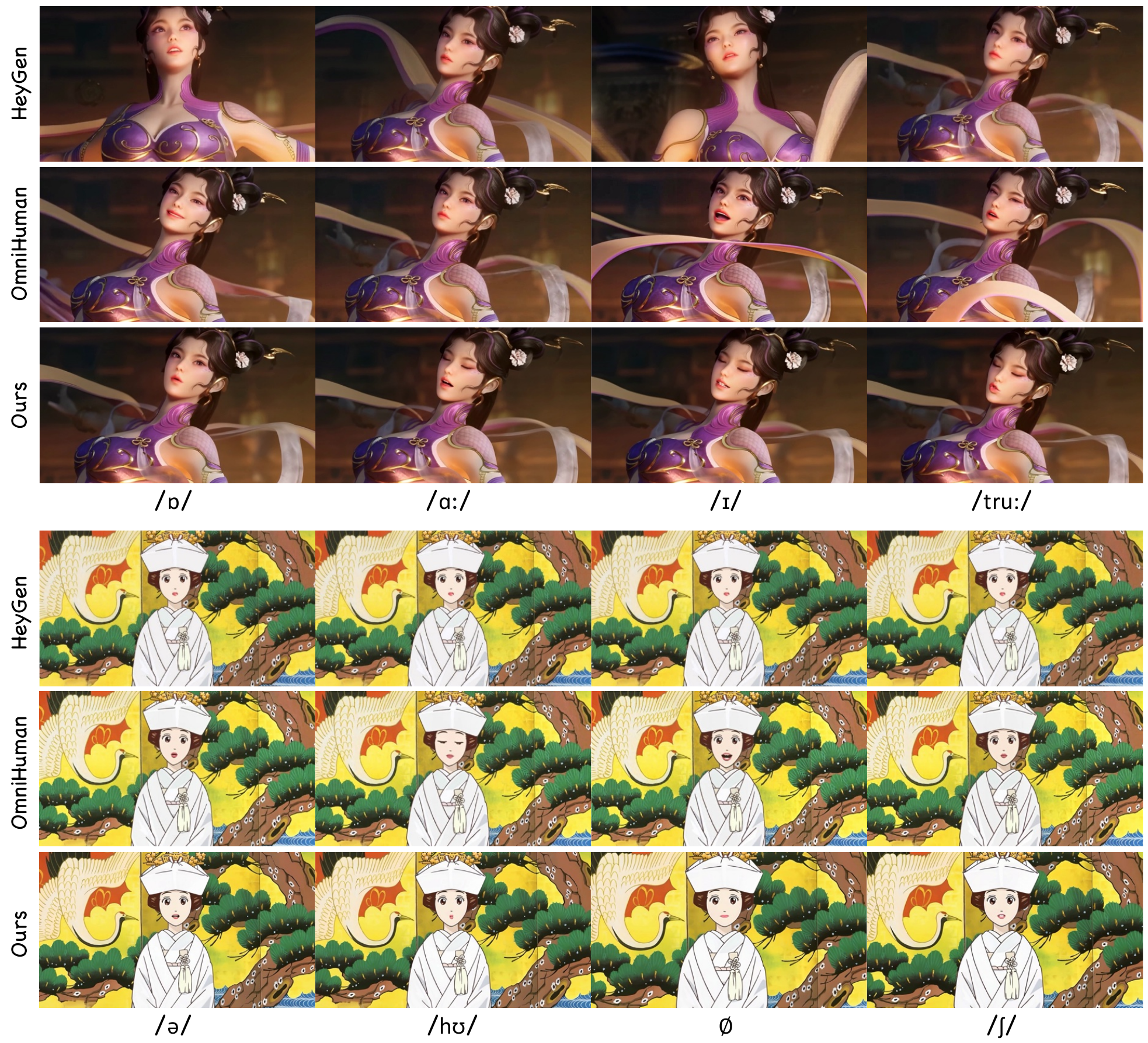}
    \vspace{-0.2cm}
    \caption{Comparison of lip synchronization between Kling-Avatar and baselines. We produce accurate lip movements for characters across different scenarios.}
    \label{fig:visual-lipsync}
\end{figure}

\clearpage
\begin{figure}[ht]
    \centering
    \includegraphics[width=\linewidth]{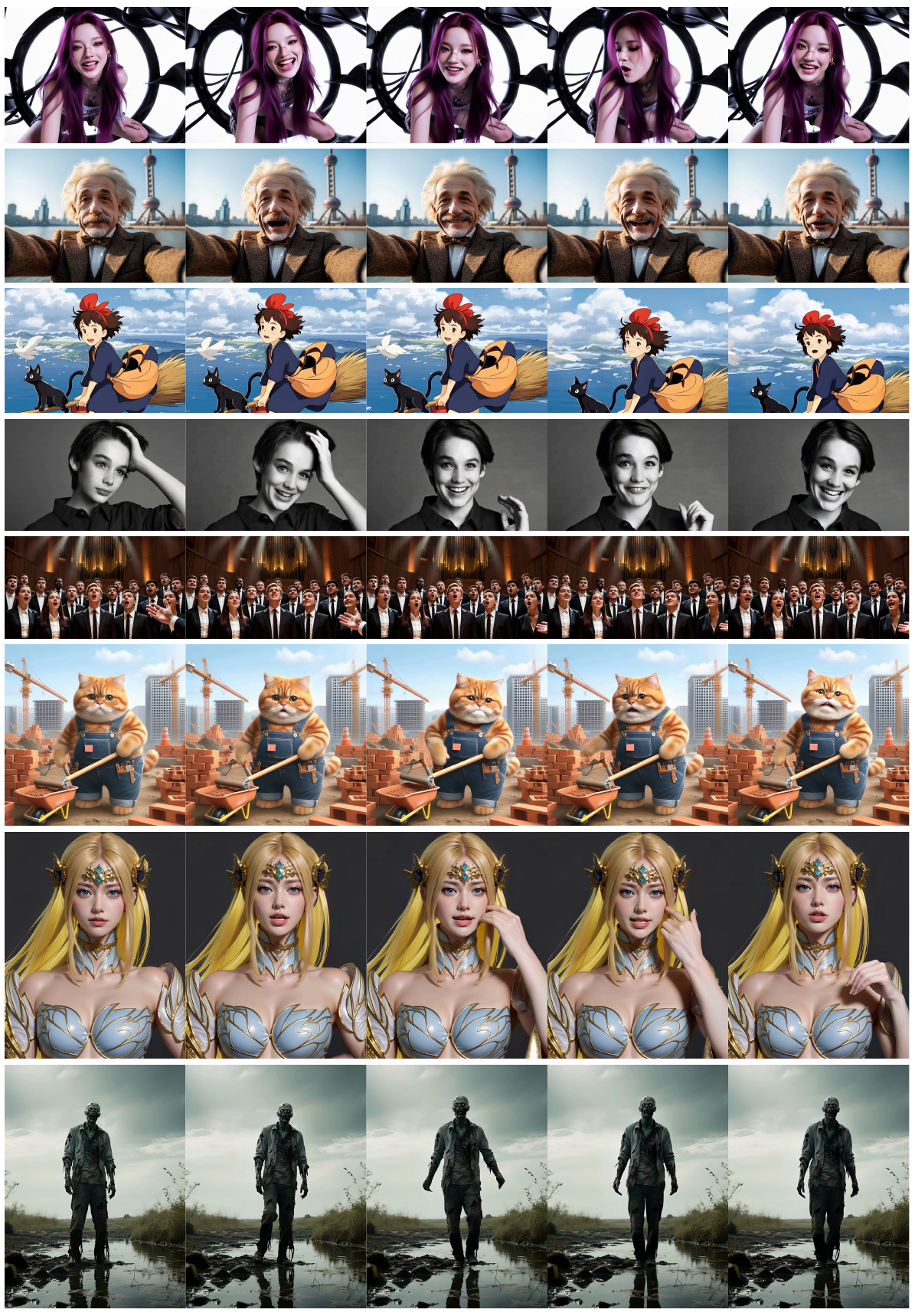}
    \vspace{-0.2cm}
    \caption{Our generated videos with multimodal instruction conditioning. We highlight our results in generating vivid and coherent portrait animations with strong control over emotions, camera movements, lip synchronization and motion dynamics.}
    \label{fig:visual-ours-short}
\end{figure}
\clearpage

\begin{figure}[t]
    \centering
    \includegraphics[width=\linewidth]{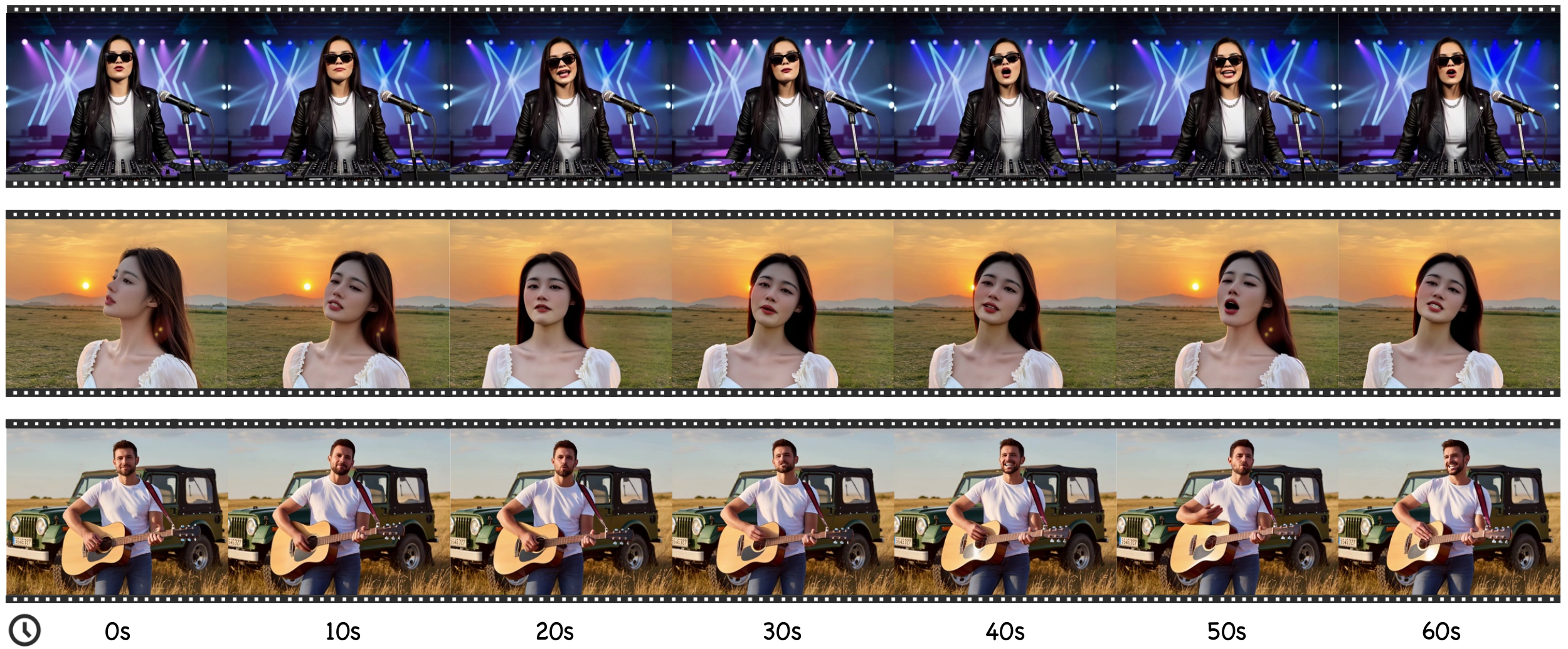}
    \vspace{-0.2cm}
    \caption{Visualization of generated long-duration videos with high consistency, coherence and vividness.}
    \label{fig:visual-ours-long}
\end{figure}

\textbf{Baselines.} We select OmniHuman-1~\citep{lin2025omnihuman1} and HeyGen~\citep{heygen} as our primary baselines, since they represent the most competitive state-of-the-art systems currently available on the market. In the future, we plan to extend our comparisons to other commercial solutions such as Higgsfield~\citep{higgsfield} and Hedra~\citep{hedra}.

\subsection{Experimental Results}

\textbf{Comparison with Baselines.} Tab.~\ref{tab:gsb} summarizes the GSB evaluation results on the benchmark, comparing our method with OmniHuman-1~\citep{lin2025omnihuman1} and HeyGen~\citep{heygen}. In addition to the overall benchmark score, we report results for three sub-categories: English speeches (Speech-En), Chinese speeches (Speech-Ch), and bilingual singing (Sing-En/Ch). Since the number of Japanese and Korean samples is relatively small, making their GSB statistics less reliable, we include them only in the overall scores. Fig.~\ref{fig:visual-exp-gsb} further visualizes the GSB comparison on the full benchmark. Numerical results show that Kling-Avatar consistently outperforms OmniHuman-1 across all dimensions, highlighting our superior performance. 

Compared with HeyGen, our method achieves notable improvements in Lip Synchronization and Visual Quality. Notably, HeyGen produces videos by repeatedly looping a five-second action pattern, which enhances motion stability and identity consistency but significantly harms vividness and diversity. In addition, HeyGen crops the reference image to fixed horizontal or vertical resolutions for generation, while our method supports arbitrary input and output resolutions, producing videos up to 1080p at 48 fps. Moreover, HeyGen is tailored for digital human scenarios, whereas our approach is developed on top of a general video generation foundation model, making it more extensible and adaptable to broader future applications. We further provide visual comparison of lip synchronization accuracy in Fig.~\ref{fig:visual-lipsync}. Our method demonstrates precise correspondence between lip shapes and various syllables, whereas the baseline methods struggle with accurate alignment and sometimes even fail to respond.

\textbf{Results on Diverse Scenarios.} Fig.~\ref{fig:visual-ours-short} showcases our diverse generation results. Benefiting from the high-level planning produced by the MLLM Director through the understanding and integration of multimodal instruction intents, our method faithfully adheres to the input signals and delivers vivid character emotions, actions, camera movements, and accurate, fine-grained lip synchronization. Moreover, it demonstrates strong generalization to various open scenarios, including multi-persons, cartoon and anime styles, and even non-human characters. Please refer to our project page for more compelling video results.

\textbf{Long-Duration Video Synthesis.} We further demonstrate the advantages of our cascaded parallel generation framework in long-duration video synthesis. As shown in Fig.~\ref{fig:visual-ours-long}, we sample one frame every 10 seconds to visualize the results. The generated frames exhibit stable identity preservation, coherent visual quality, and rich character dynamics. Notable examples include the background lighting changes in the first line, the head movements in the second, and the hand gestures in the third.

\section{Related Work}
\subsection{Video Generation}
Breakthroughs in diffusion models for image synthesis~\citep{ho2020ddpm,dhariwal2021diffusion,rombach2022high} have driven an evolution in video generation, where scalable training paradigms based on noise inversion and conditional denoising have made high-fidelity appearance and controllability attainable. Early video generation approaches typically extend pretrained image-based U-Nets by stacking temporal modules or inserting temporal attention into spatial backbones to capture cross-frame relations~\citep{ho2022video,singer2022make,blattmann2023stable}. However, such designs face limitations in scalability in terms of resolution and sequence length. Recently, the growth of training data and computational resources has shifted the focus toward Diffusion Transformers (DiT)~\citep{peebles2023scalable,yang2025cogvideox,wan2025wan}. This paradigm compresses videos into spatiotemporal tokens via 3D VAEs, and leverages the large context capacity and scalable attention of transformers to capture temporal dynamics, thus supporting stable large-scale video generation and establishing itself as the emerging mainstream approach. It also demonstrates strong potential for long-term generation~\citep{kong2024hunyuanvideo,zhang2025packing,teng2025magi,chen2025midas}, real-time synthesis~\citep{zhao2025real,gu2025far}, and world modeling~\citep{yu2025context,team2025hunyuanworld}. These methods are primarily designed for general video generation yet remain inadequate for speech-driven digital portrait modeling.

\subsection{Audio-Driven Digital Human Synthesis}
Audio-driven digital human synthesis aims to generate realistic and expressive talking videos conditioned on speech signals and a reference portrait image. One line of work employs explicit intermediate representations, such as facial landmarks or 3D head models, to drive facial expressions and lip movements~\citep{chen2025cafe,hu2025ggtalker,guo2024liveportrait,cui2025cfsynthesis}. However, these approaches are typically limited to facial animation and cannot produce natural upper-body motion or hand gestures. More recent studies leverage diffusion models for end-to-end audio-driven video generation. By directly injecting speech as a condition into diffusion transformers, these methods achieve joint alignment and control of audio, expressions, and motion within a unified attention framework~\citep{jiang2025omnihuman1.5,gan2025omniavatar,peng2025omnisync,wang2025fantasytalking,guo2024liveportrait}, enabling realistic and coherent video synthesis without relying on 3D priors, and showing advantages in expression detail and lip–audio alignment. To further support hand motion and human–object interactions, methods such as Emo2~\citep{tian2025emo2} and HunyuanVideo-HOMA~\citep{huang2025hunyuanhoma} incorporate pose sequences as conditions alongside speech and body dynamics. Other approaches such as Mocha~\citep{wei2025mocha}, MultiTalk~\citep{kong2025multitalk} and InteractHuman~\citep{wang2025interacthuman}, learn identity information or memory-slot IDs to enable speaker switching and cross-shot localization. Additional efforts explore data scaling~\citep{lin2025omnihuman1}, audio–video alignment strategies~\citep{gao2025wans2v}, and direct performance optimization~\citep{cui2025hallo4}. Despite these advances, existing methods still rely on local cues for alignment within each modality, and thus struggle with multimodal instruction understanding and consistent long-duration generation. To address these challenges, we explore the use of multimodal large language models for instruction grounding and propose a cascaded framework for fast synthesizing vivid, long-duration portrait animations.

\section{Conclusion}

In this paper, we introduce Kling-Avatar, a cascaded framework that unifies multimodal instruction understanding with long-duration generation of lifelike portrait videos. Our two-stage pipeline first employs an MLLM director to produce a blueprint video that encodes high-level semantic intent into a coherent storyline, and then synthesizes long videos through parallel sub-clip generation guided by blueprint keyframes to refine local dynamics. Coupled with carefully curated data and practical training and inference strategies, our framework preserves fine-grained details while faithfully realizing global semantics. To evaluate the effectiveness, we construct a 375-sample benchmark spanning diverse instructions and challenging scenarios. Experiments demonstrate that Kling-Avatar delivers vivid, fluent videos up to 1080p and 48 fps, with precise lip synchronization, strong controllability, and robust generalization to open scenarios. Human preference–based metric comparisons further confirm our superior performance. We believe our exploration of instruction-grounded, long-duration avatar video generation represents a promising step toward broad real-world applications and future research.

\clearpage

\bibliography{colm2024_conference}
\bibliographystyle{colm2024_conference}

\end{document}